\documentclass[sn-nature]{sn-jnl}% Style for submissions to Nature Portfolio journals
\usepackage{graphicx}%
\usepackage{multirow}%
\usepackage{amsmath,amssymb,amsfonts}%
\usepackage{amsthm}%
\usepackage{mathrsfs}%
\usepackage[title]{appendix}%
\usepackage{xcolor}%
\usepackage{textcomp}%
\usepackage{manyfoot}%
\usepackage{booktabs}%
\usepackage{algorithm}%
\usepackage{algorithmicx}%
\usepackage{algpseudocode}%
\usepackage{placeins}
\usepackage{listings}%
\theoremstyle{thmstyleone}%
\theoremstyle{thmstyletwo}%

\theoremstyle{thmstylethree}%

\begin{document}

% \title[Article Title]{XiChen: Observation-scalable assimilation for data-driven global weather forecasting using variational gradients as unified features}
\title[Article Title]{XiChen: A global weather observation-to-forecast machine learning system via four-dimensional variational gradient-guided flexible assimilation}
% \tilde[Article Title]{A variational gradient-guided machine learning framework for global weather forecasting from heterogeneous observations}
% \title[Article Title]{A four-dimensionala variational guided global weather data-to-forecast machine learning system from heterogeneous and evolving observtions}

\author[1,2]{\fnm{Wuxin} \sur{Wang}}\email{wuxinwang@nudt.edu.cn}
 
\author[2]{\fnm{Weicheng} \sur{Ni}}\email{niweicheng17@nudt.edu.cn}

\author*[3]{\fnm{Ben} \sur{Fei}}\email{benfei@cuhk.edu.hk}

\author[3]{\fnm{Tao} \sur{Han}}\email{hantao10200@gmail.com}

\author[1,2]{\fnm{Lilan} \sur{Huang}}\email{huanglilan18@nudt.edu.cn}

\author[2]{\fnm{Shuo} \sur{Ma}}\email{mashuo@nudt.edu.cn}

\author[2]{\fnm{Taikang} \sur{Yuan}}\email{ytk@nudt.edu.cn}

\author[2]{\fnm{Yanlai} \sur{Zhao}}\email{zhaoyanlai@nudt.edu.cn}

\author[2]{\fnm{Kefeng} \sur{Deng}}\email{dengkefeng@nudt.edu.cn}

\author[2]{\fnm{Xiaoyong} \sur{Li}}\email{sayingxmu@nudt.edu.cn}

\author[2]{\fnm{Hongze} \sur{Leng}}\email{hzleng@nudt.edu.cn}

\author*[2]{\fnm{Boheng} \sur{Duan}}\email{bhduan@nudt.edu.cn}

\author*[3]{\fnm{Lei} \sur{Bai}}\email{bailei@pjlab.org.cn}

\author[1,2]{\fnm{Weimin} \sur{Zhang}}\email{weiminzhang@nudt.edu.cn}

\author[1,2]{\fnm{Junqiang} \sur{Song}}\email{junqiang@nudt.edu.cn}
	
\author*[1,2]{\fnm{Kaijun} \sur{Ren}}\email{renkaijun@nudt.edu.cn}

\affil*[1]{\orgdiv{College of Computer Science and Technology}, \orgname{National University of Defense Technology}, \orgaddress{\street{Deya Street}, \city{Changsha}, \postcode{410073}, \state{Hunan}, \country{China}}}
      
\affil*[2]{\orgdiv{College of Meteorology and Oceanography}, \orgname{National University of Defense Technology}, \orgaddress{\street{Deya Street}, \city{Changsha}, \postcode{410073}, \state{Hunan}, \country{China}}}

\affil*[3]{\orgdiv{Shanghai Artificial Intelligence Laboratory}, \orgaddress{\postcode{200030}, \state{Shanghai}, \country{China}}}


\abstract{
Machine Learning (ML) has shown great promise in revolutionizing weather forecasting, yet most ML systems still rely on initial conditions generated by Numerical Weather Prediction (NWP) systems. End-to-end ML models aim to eliminate this dependency, but they often rely on observation-specific encoders and require redesign or retraining when observation sources change, thereby limiting their operational robustness. Here, we introduce XiChen, a global weather observation-to-forecast ML system via four-dimensional variational (4DVar) gradient-guided flexible assimilation. We demonstrate that the gradient of the 4DVar cost function serves as a physically grounded interface that maps heterogeneous observations into a common state space. This novel formulation enables XiChen to flexibly assimilate diverse conventional and raw satellite observations while preserving physical consistency. Experiments show that the system achieves forecasting metrics competitive with operational NWP systems. This work provides a practical and physically consistent route toward operational ML-based global weather forecasting systems with heterogeneous and evolving observations.
}




\maketitle

\section{Introduction}\label{sec1}
Accurate medium-range global weather forecasting is a cornerstone for transportation, disaster early warning systems, and renewable energy integration. Over the past decades, physics-based Numerical Weather Prediction (NWP) systems have achieved remarkable progress~\cite{bauer2015quiet}, driven by synergistic advances in computational power~\cite{bi2023accurate}, Data Assimilation (DA) techniques~\cite{wang2000data,bannister2017review}, observational infrastructures~\cite{tatem2008fifty}, and physical understanding~\cite{gettelman2022future}. However, as we enter an era of explosive growth in global satellite and sensor data streams, traditional NWP systems are approaching practical computational limits~\cite{bauer2021digital,ecmwf10years}. This limitation suggests that incremental improvements to existing NWP pipelines are insufficient to keep pace with the influx of data from future observational systems. Consequently, there is a compelling motivation to transition toward fundamentally new paradigms that can fully exploit data-rich environments.

Recent advances in Machine Learning (ML) have demonstrated substantial potential for simulating complex weather systems without the costly step-by-step numerical integration of partial differential equations~\cite{bi2023accurate,lam2023learning,fudanchen2023fuxi,ailabchen2025fengwu}. However, most ML-based weather forecasting models lack inherent DA capabilities and typically rely on initial conditions provided by traditional NWP systems~\cite{chen2023fengwuadas,xiao2024fengwu4dvar,wang2024accurate}, preventing the realization of a fully self-contained pipeline from observations to forecasting results. Emerging end-to-end ML models~\cite{chen2023fengwuadas,xu2024fuxida,mcnally2024data,alexe2024graphdop,allen2025end,sun2025data} seek to overcome this dependency, but they face a major limitation in adapting to the explosive growth and extreme heterogeneity of global satellite and sensor data~\cite{gettelman2022future,bauer2021digital}. Current approaches rely on observation-type-specific encoders tightly coupled to sensor characteristics, requiring redesign or retraining the entire pipeline whenever observation sources are missing, evolved, or are newly introduced~\cite{xu2024fuxida,allen2025end,sun2025data}. This limitation is particularly problematic given the rapid evolution of Earth observation systems, where new satellite platforms and sensing modalities are continuously deployed~\cite{gettelman2022future}. Consequently, the central challenge in achieving a fully self-contained ML pipeline is the absence of a physically consistent and flexible ML-based DA framework that scales with evolving global observations.

Integrating the four-dimensional variational (4DVar)~\cite{carrassi2018data} DA method with ML techniques offers a promising approach to addressing the above challenge. This synergy is driven by the fact that the gradient of the 4DVar cost function encodes the sensitivity of atmospheric states to observational discrepancies, reflecting the physical constraints imposed by observations on the initial state through the forecasting model. Recent ML-based 4DVar frameworks leverage the automatic differentiation capabilities of ML-based weather forecasting models to replace the tangent-linear and adjoint models traditionally required in operational 4DVar systems~\cite{xiao2024fengwu4dvar,li2024fuxien4dvar,wang2024accurate}. However, existing research has primarily utilized simulated conventional observations and has yet to demonstrate scalability to real-world, high-dimensional, and heterogeneous observations. To bridge this gap, it is essential to internalize the 4DVar constraint within ML architectures. This integration can facilitate a unified and physically consistent framework capable of learning complex observation-to-state mappings directly from diverse conventional and raw satellite observations.

Here, we propose XiChen, a flexible ML-based global DA and medium-range global weather forecasting system. To address the challenges posed by data heterogeneity, XiChen leverages a foundation model pre-trained for global weather forecasting, which is subsequently fine-tuned to function as observation operators and DA models. By utilizing the 4DVar gradient as a physically consistent unifying information interface, XiChen transforms multi-source observations into a common state space that the DA model can seamlessly ingest. Within this framework, the 4DVar cost function is formulated using the ML-based weather forecasting model and observation operators, with its gradient relative to the background field computed via automatic differentiation. This approach enables XiChen to facilitate stable, long-term DA cycles by leveraging multi-source heterogeneous real-world observations. Experimental results indicate that the DA process of XiChen achieves physical consistency and flow-dependent behavior, reflecting the core advantages of integrating 4DVar within the ML model. Furthermore, the medium-range global weather forecasting metrics of XiChen are comparable to those of advanced operational NWP systems for key atmospheric variables. Finally, we demonstrate its potential for the continuous, scalable assimilation of heterogeneous observations and the iterative improvement of forecasting metrics.

\section{Results}\label{sec2}
\begin{figure}[htb]%
    \centering
    \includegraphics[width=0.95\textwidth]{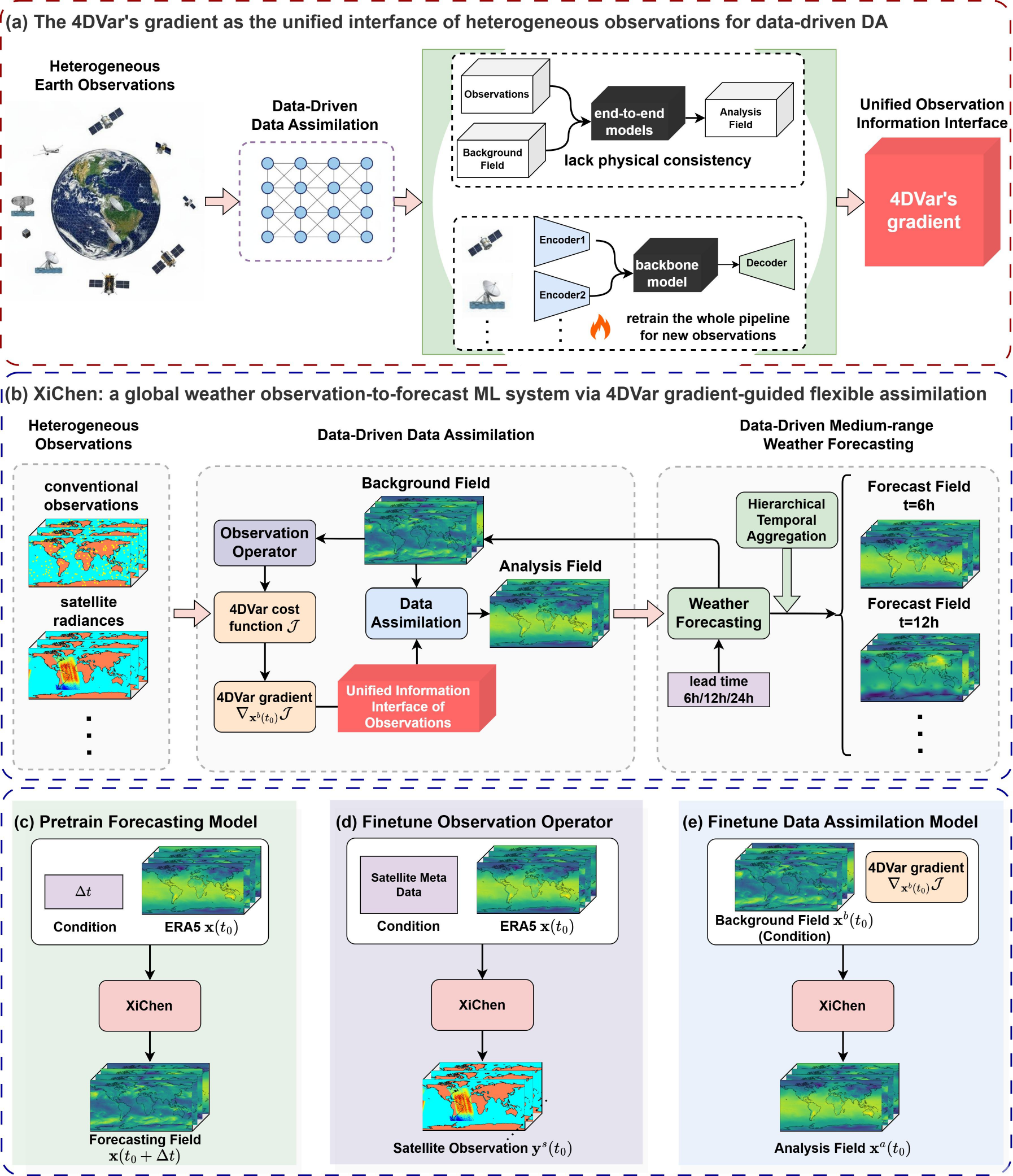}
    \caption{\textbf{Overview of our XiChen.} (a) A schematic illustration demonstrating the utility of the 4DVar cost function gradient as a unified information interface of observations. (b) The overall framework of ``XiChen'': a global weather observation-to-forecast machine ML system via 4DVar gradient-guided flexible assimilation. XiChen integrates background field data, conventional observations, and indirect satellite observations within the Data Assimilation Window (DAW) to perform the DA task. The process begins by computing the 4DVar cost function, and it subsequently determines the gradient with respect to the background field. By combining this gradient with the background field, XiChen generates the analysis field. Utilizing the analysis field along with the lead time as inputs, XiChen produces either a forecast field or an updated background field. The green box denotes the weather forecasting model, the purple represents the observation operator, and the blue signifies the DA model. XiChen undergoes three stages of training: (c) It is pre-trained as a medium-range global weather forecasting model, with lead time ($\Delta t$) as the conditional input; (d) It is fine-tuned as an observation operator, using meta data of the satellite as the conditional input; (e) It is fine-tuned as a DA model, with the gradient of the 4DVar cost function as the input and the background field as the conditional input.}
\label{fig1}
\end{figure}

As illustrated in Figure~\ref{fig1}(a), existing data-driven DA methods encounter two primary challenges in handling the complex heterogeneous observation system: (1) end-to-end models fuse observations and background fields in a purely black-box manner, making it difficult to ensure physical consistency; and (2) each observation type typically requires a specific encoder, necessitating a full pipeline retraining to accommodate missing, evolving, or newly introduced observations. To address these limitations, we show that the gradient of the 4DVar cost function serves as a unified information interface for heterogeneous observations. This principle enables XiChen, a physically consistent ML-based system for medium-range global weather forecasting and observation-flexible DA. As illustrated in Figure~\ref{fig1}(b) and (c), we implement XiChen by pretraining a foundation model as a medium-range global weather forecasting model and subsequently fine-tuning it as the observation operators and DA models. Specifically, we construct a 4DVar cost function by integrating the forecasting model with the observation operators. The gradient of the 4DVar cost function is then calculated via automatic differentiation to propagate information from heterogeneous observations across the spatial and multi-variable dimensions of the atmospheric state (see Section~\ref{sec-method} for details). For details of the DA cycle and medium-range forecasting experiments, see the Supplementary Text, Figure S1(a) and S1(b).

XiChen is evaluated over one year in 2023. The evaluation focuses on four key aspects: (1) single-point perturbation assimilation of satellite radiances to assess the physical consistency of the DA process of XiChen; (2) the skill of XiChen's 10-day medium-range global weather forecasts; (3) the impact of assimilating different observations on medium-range global weather forecasts; and (4) the influence of satellite assimilation on Tropical Cyclone (TC) forecasting skill. The DA process employs a cascade framework, sequentially assimilating conventional observations from the Global Data Assimilation System (GDAS) prepbufr data, followed by Microwave Humidity Sensor (MHS) brightness, wind field retrievals from the Advanced Scatterometer (ASCAT), Advanced Microwave Sounding Unit-A (AMSU-A) brightness, and satellite-derived winds (SATWND) observations. The weather forecasting model, observation operators, and DA models share the same underlying model architecture and are trained using different data. For medium-range global weather forecasting, we adopted configurations and evaluation metrics from WeatherBench~\cite{rasp2020weatherbench} and DABench~\cite{wang2024benchmark} to generate initial fields and evaluate XiChen’s 10-day forecasts (see Section~\ref{sec-metrics} and Figure S1(c) for details).

\FloatBarrier

\subsection{Physical consistency of the assimilation}

To evaluate whether the XiChen assimilation process exhibits fundamental physical consistency, we designed a single-point perturbation experiment to assess the system's response to localized observation changes. The evaluation consists of two separate DA runs. The first run used ERA5 as the background field to assimilate raw satellite observations. In the second run, a perturbation is introduced to the raw satellite inputs across an individual observation channel. The differences between the two DA results illustrate the impact of the perturbation on the analysis fields, revealing how observational information propagates through the variables and spatial dimensions. The DA time was set to 00 UTC on July 24, 2023, and perturbation experiments were conducted separately for AMSU-A and MHS. The perturbation location, selected at 20°N, 120°E, was near the path of Typhoon Doksuri during this period. The perturbation magnitude was set to +5 K, and independent experiments were performed for each satellite channel.

\begin{figure}[htb]%
    \centering
    \includegraphics[width=0.95\textwidth]{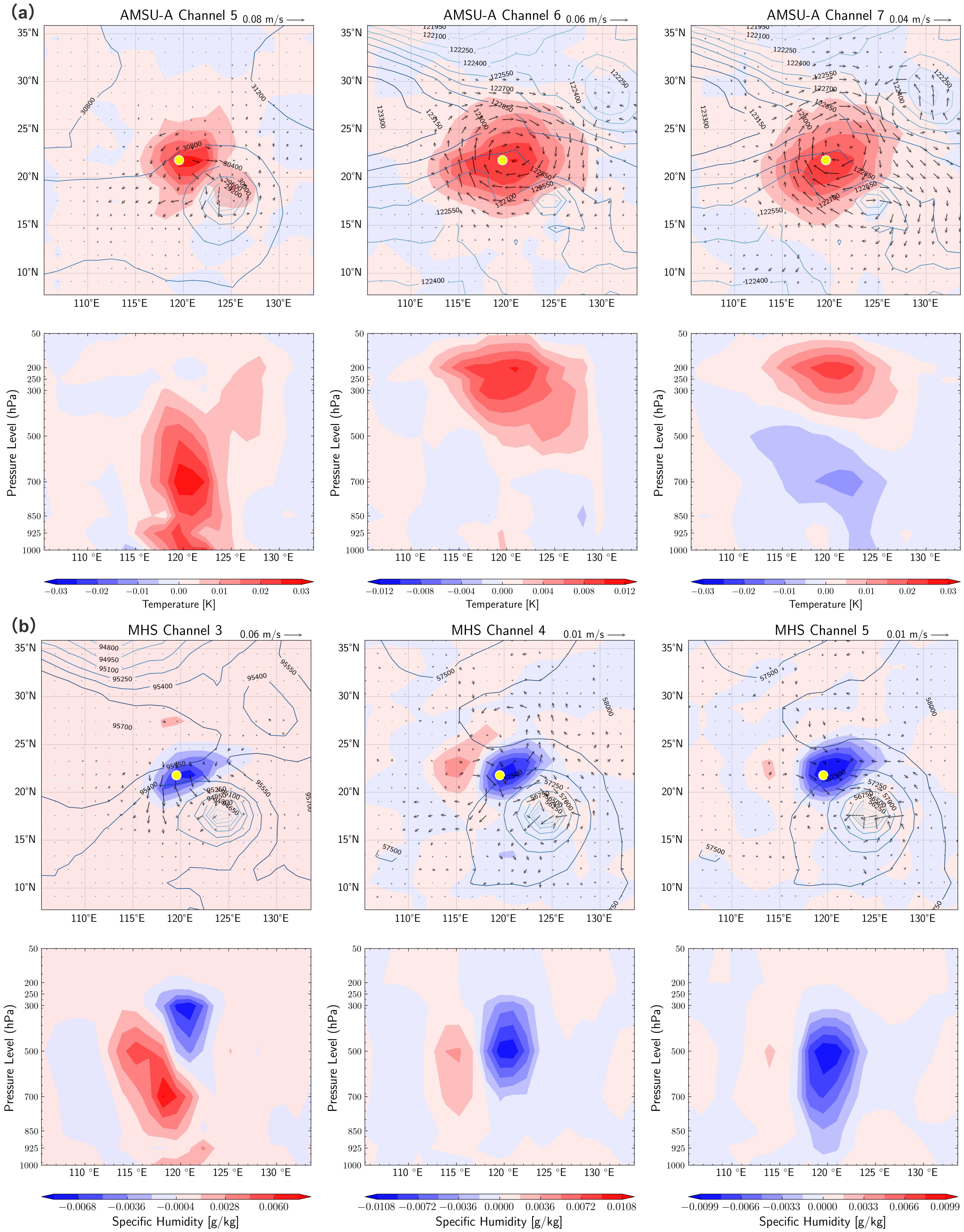}
    \caption{\textbf{Analysis field changes from a 5 K perturbation to AMSU-A and MHS observations.} The perturbation was applied at 03 UTC near Typhoon Doksuri (20$^\circ$N, 120$^\circ$E; yellow dot). (a) AMSU-A results (channels 5-7). Top row: horizontal temperature changes (shading, K) and wind differences (vectors, m s$^{-1}$) at 700, 200, and 200 hPa. Solid contours show background geopotential (m$^2$ s$^{-2}$). Bottom row: vertical cross-section. (b) MHS results (channels 3-5). Top row: horizontal specific humidity changes (shading, g kg$^{-1}$) and wind differences (vectors, m s$^{-1}$) at 300, 500, and 500 hPa. Bottom row: vertical cross-section.}
\label{fig2}
\end{figure}

Figures~\ref{fig2}(a) and \ref{fig2}(b) present the horizontal and vertical distributions of analysis field variations resulting from individually perturbing three AMSU-A channels (5, 6, and 7) and three MHS channels (3, 4, and 5), respectively. For AMSU-A, increases in brightness temperature correspond to rising atmospheric temperatures, consistent with radiative transfer theory, demonstrating that XiChen has effectively learned AMSU-A’s temperature sensitivity. For MHS, increases in brightness temperature correspond to decreases in specific humidity, indicating that XiChen has similarly captured MHS’s sensitivity to atmospheric moisture. In both experiments, the analysis increments exhibit propagating variations horizontally and vertically and display localized, anisotropic structures around the observations that align with the background geopotential field, reflecting the flow-dependent nature of the XiChen system. Additionally, for AMSU-A, anticyclonic wind responses surrounding temperature increments imply that XiChen satisfies certain equilibrium constraints. Overall, these results demonstrate that XiChen adheres to fundamental physical consistency, which is consistent with the traditional 4DVar system~\cite{zhu2022four}.

\FloatBarrier

\subsection{Medium-range global weather forecasting}
As forecasting accuracy is the primary criterion for the ML-based global DA and medium-range global weather forecasting system, we evaluate the performance of XiChen against advanced operational NWP systems. This evaluation is based on global weather forecasts at 6-hour intervals. The temporal initialization of these analysis fields aligns with the configurations established by WeatherBench~\cite{rasp2020weatherbench} and DABench~\cite{wang2024benchmark}.

\begin{figure}[htb]%
    \centering
    \includegraphics[width=0.95\textwidth]{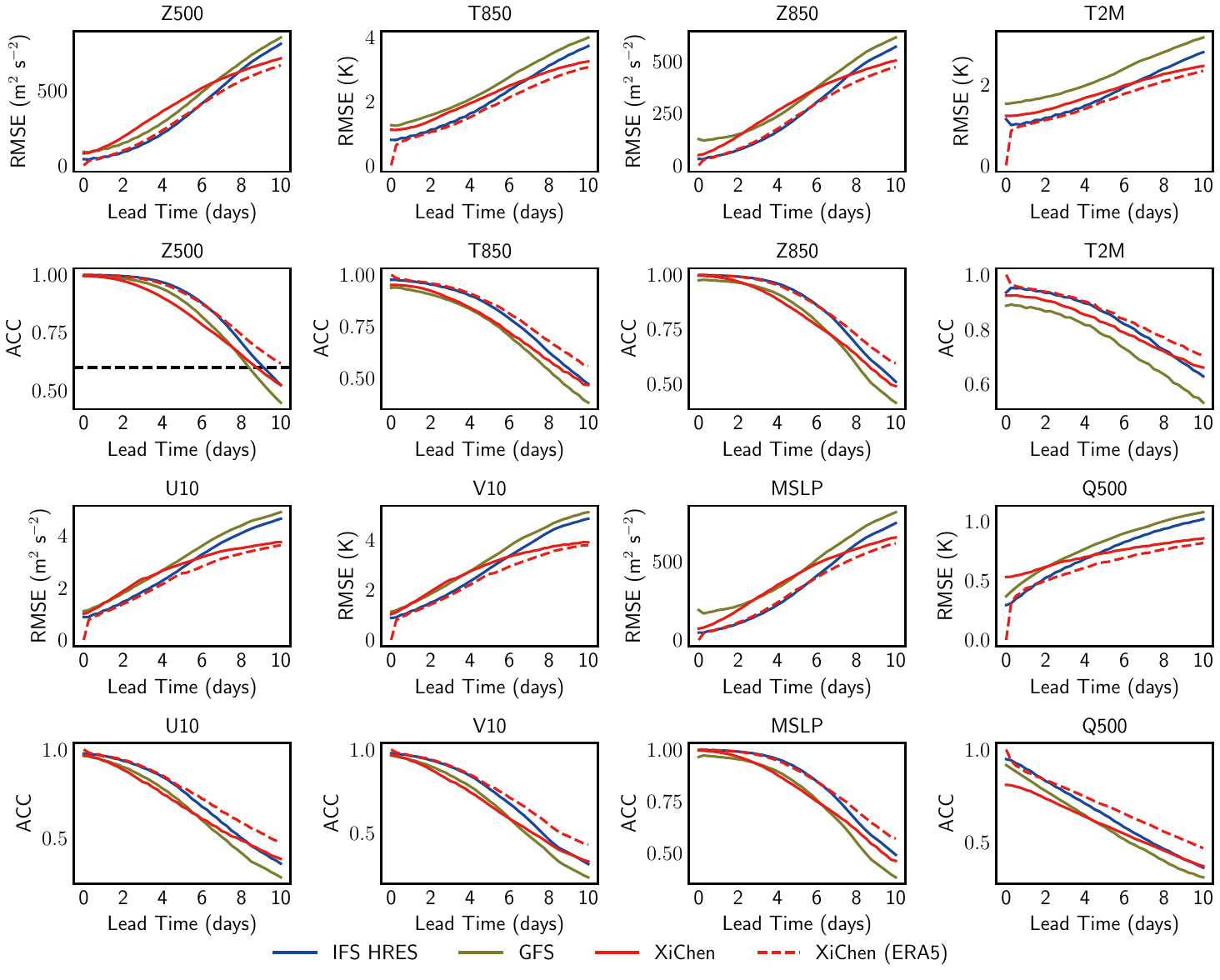}
    \caption{\textbf{Comparison of the average latitude RMSE (first and third rows) and ACC (second and fourth rows) of the 10-day medium-range global weather forecasting using XiChen (red lines) as well as the operational forecasting results of GFS (yellow lines) and IFS HRES (blue lines) using testing data during 2023. The evaluation encompasses eight meteorological variables: geopotential height at 500 hPa (Z500) and 850 hPa (Z850), 2-meter temperature (T2M), temperature at 850 hPa (T850), Mean Sea Level Pressure (MSLP), 10-meter zonal (10U) and meridional (10V) wind components, as well as specific humidity at 500 hPa (Q500).} All analysis fields are evaluated against the ERA5 reanalysis dataset.}
\label{fig3}
\end{figure}

Figure~\ref{fig3} illustrates the averaged latitude-weighted Root Mean Square Error (RMSE) and Anomaly Correlation Coefficient (ACC), both as a function of forecast lead times over 10 days. XiChen achieves a skillful weather forecasting lead time exceeding 10.25 days (with ACC of Z500 $> 0.6$). This performance rivals advanced medium-range global weather forecasting models trained on 0.25$^\circ$ high-resolution datasets—Pangu~\cite{bi2023accurate}, GraphCast~\cite{lam2023learning}, FengWu~\cite{ailabchen2025fengwu}, and FuXi~\cite{fudanchen2023fuxi}.

\FloatBarrier

When using XiChen's self-generated initial conditions, XiChen achieves a skillful weather forecasting lead time exceeding 8.75 days, surpassing the GFS system (8.25 days). Notably, for key variables such as T2M, T850, MSLP, U10, and V10, the global weather forecasts of XiChen deliver lower RMSEs compared to the GFS for nearly the entire 10-day lead time. The RMSEs for all variables approach or even surpass the performance of the IFS HRES as the lead time increases. For RMSE of Q500, XiChen exceeds the GFS after 2 days and the IFS HRES after 4 days. Although the forecast error for Z500 remains higher than that of both the GFS and the IFS HRES during the first 6 days, XiChen surpasses the GFS after 7 days and outperforms IFS HRES after 8 days. This achievement is particularly noteworthy considering that XiChen was trained using only 12 years of ERA5 data at a 1.40625$^\circ$ resolution. Moreover, XiChen assimilated a limited subset of observations used in operational NWP systems, including GDAS prepbufr, MHS, ASCAT, AMSU-A, and SATWND, significantly fewer than those used by GFS and IFS HRES. XiChen's RMSE surpasses that of IFS HRES at larger lead times, which may be because its forecasts tend to favor smoothing, as evidenced by the power spectra presented in Figure S6. This tendency is a common characteristic observed in contemporary ML-based weather forecasting models~\cite{lam2023learning,ailabchen2025fengwu,fudanchen2023fuxi}. For comprehensive insights into the 10-day medium-range forecast RMSE scorecards across all variables for XiChen versus GFS, and XiChen versus IFS HRES, kindly consult the Supplementary Text, as well as Figures S7 and S8. Furthermore, we present visualizations of the forecasting results produced by XiChen, IFS HRES, and GFS in Figures S12 to S17.

\subsection{The impact of assimilating different observations on medium-range global weather forecasts}
A pivotal innovation of XiChen lies in its scalable and flexible capability to integrate heterogeneous observations within a cascade DA framework for the accurate estimation of initial conditions (see Figure S1 for details). With the continuous expansion in both the number and diversity of observational modalities, a pressing question emerges: which specific observations exert the most significant influence on medium-range global weather forecasting accuracy~\cite{laloyaux2025using}. To answer this question, we designed an ablation experiment aimed at quantifying the marginal impact of omitting individual observation sources on 10-day medium-range forecast errors.

\begin{figure}[htb]%
    \centering
    \includegraphics[width=\textwidth]{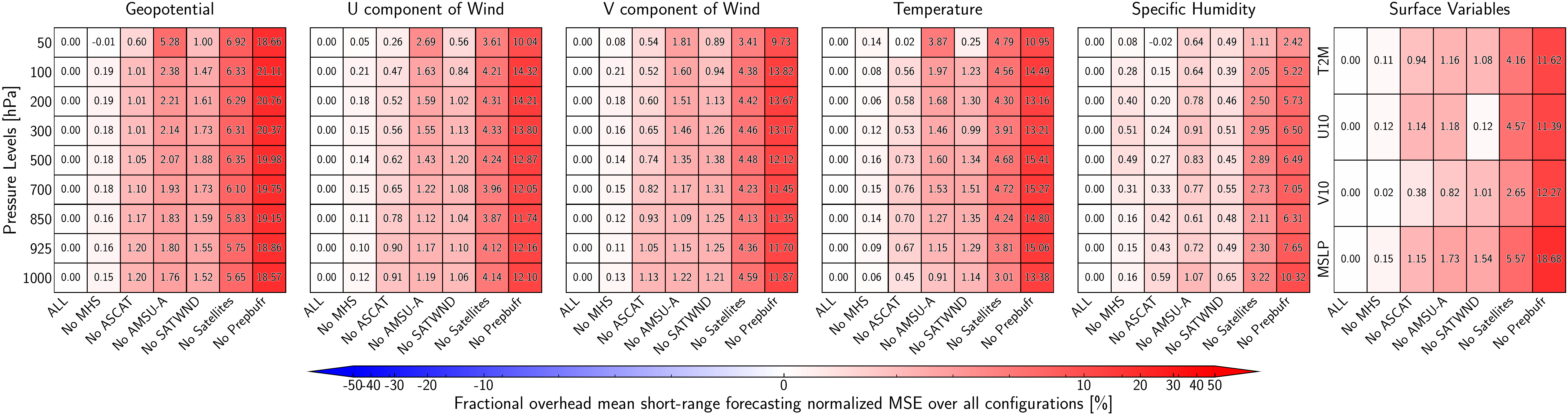}
    \caption{\textbf{Headline scorecards of the average normalized RMSEs for 10-day medium-range global weather forecasting, based on the assimilation of different observations.} The colors denote the percentage difference relative to the baseline, which assimilates all observations, including GDAS prepbufr, AMSU-A, MHS, ASCAT, and SATWND. Blue shading indicates a reduction in normalized RMSE, while red shading indicates an increase.}
\label{fig4}
\end{figure}

In this experiment, various observation sources were systematically excluded from the assimilated dataset, after which DA was conducted using XiChen. The resulting 10-day forecast mean errors were subsequently compared against those derived from assimilating the complete set of available observations (designated as the ``All" configuration). For instance, configurations such as ``No ASCAT" and ``No Satellites" denote the exclusion of ASCAT data and all satellite-derived observations, respectively, from the ``All" configuration. Figure~\ref{fig4} provides a detailed depiction of the percentage increase in the 10-day mean latitude-weighted RMSE across all test cases, benchmarked against the ``All" configuration.

The results reveal that removing any observation source increases errors in medium-range global weather forecasts. Among these, GDAS prepbufr has the most significant impact, likely due to its direct measurements of fundamental atmospheric variables. Furthermore, removing ASCAT or SATWND increases errors across all variables. Since ASCAT and SATWND observe surface winds and upper-air winds, respectively, this highlights XiChen's effectiveness in learning correlations among atmospheric variables. This capability enables the system to use observed variables to correct unobserved ones. 

\FloatBarrier

Removing data from channels 5 to 10 of the AMSU-A satellite, which are primarily used to observe atmospheric temperature profiles, resulted in increased errors in upper-air temperatures~\cite{wang2014amsu}. This suggests that the observation operator in XiChen effectively captures AMSU-A's sensitivity to atmospheric temperature profiles. Moreover, excluding AMSU-A observations increased errors in the upper-air geopotential fields and introduced errors in the wind fields. These effects likely arise from XiChen’s ability to implicitly adjust the pressure gradient force in response to temperature variations, thereby influencing the potential and wind fields. Additionally, the exclusion of AMSU-A data led to increased humidity errors, highlighting XiChen’s ability to learn the temperature-humidity balance relationship.

Similarly, the exclusion of data from channels 3 to 5 of the MHS, which are predominantly utilized for observing atmospheric humidity profiles~\cite{zou2017impacts}, led to increased errors in upper-air specific humidity. This demonstrates the effectiveness of the observation operator within the XiChen framework in accurately encapsulating MHS's sensitivity to atmospheric humidity profiles. Moreover, the omission of MHS observations was observed to introduce elevated errors across other variables to a certain degree. This observation further emphasizes XiChen's capability to leverage its learned equilibrium constraints, effectively redistributing the specific humidity adjustments to other related variables within the system.

\begin{figure}[htb]%
    \centering
    \includegraphics[width=\textwidth]{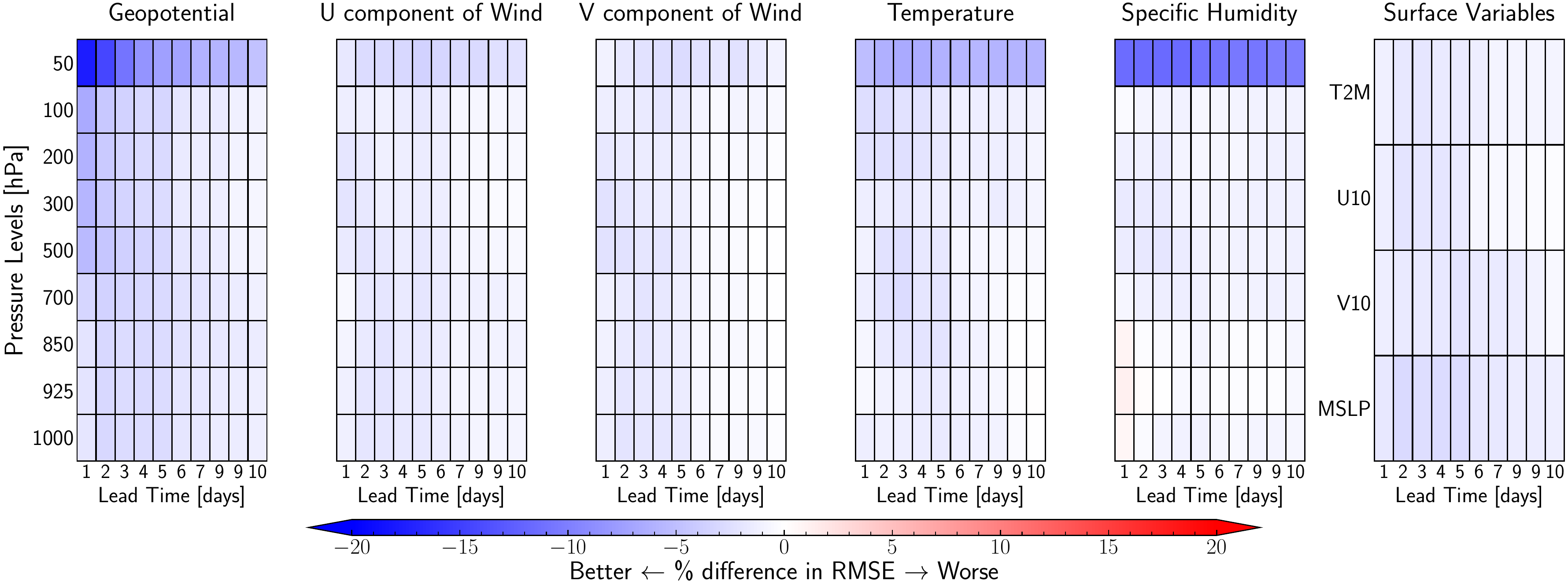}
    \caption{\textbf{Headline scorecards of the global-averaged latitude-weighted RMSEs for 10-day medium-range global weather forecasting, based on the assimilation of different observations.} The 10-day medium-range forecasts based on analysis fields that assimilate all observations are compared with the forecasts based on analysis fields that assimilate only the GDAS prepbufr. The colors indicate the percentage difference relative to the baseline, which assimilates only the GDAS prepbufr. Blue shading represents a reduction in RMSE, while red shading represents an increase.}
\label{fig5}
\end{figure}
\FloatBarrier

Notably, the exclusion of all satellite observations results in a substantially greater increase in forecast errors compared to the removal of a single satellite observation type. This highlights the robust capability of XiChen to effectively integrate a diverse array of satellite observations, thereby enhancing forecast accuracy. Additionally, Figure~\ref{fig5} illustrates the influence of additionally assimilating data from MHS, ASCAT, AMSU-A, and SATWND, alongside the GDAS prepbufr, on the RMSE of 10-day medium-range global weather forecasts. Across nearly the entire forecast horizon, the incorporation of satellite observations markedly reduces forecast errors for all variables. This reduction is particularly pronounced for upper-atmospheric variables, such as geopotential, temperature, and specific humidity, underscoring the critical role of satellite data in improving the accuracy of medium-range global weather predictions.

\subsection{Assimilation and forecasting of tropical cyclones}
To assess XiChen’s ability to assimilate and forecast TC trajectories, we employ the TempestExtremes tracker~\cite{ullrich2021tempestextremes} to identify TC tracks. For comparison with the IBTrACS~\cite{Kenneth2010IBTrACS} dataset and ERA5 reanalysis, we adopt the configuration used in NeuralGCM~\cite{kochkov2024neural}, with parameter details listed in Table S9. To maintain consistency with IBTrACS, we exclude any TCs identified by this configuration that are not present in the IBTrACS dataset, ensuring that only IBTrACS TCs are included in our results (see the Supplementary Text for details).

\begin{figure}[htb]%
    \centering
    \includegraphics[width=0.95\textwidth]{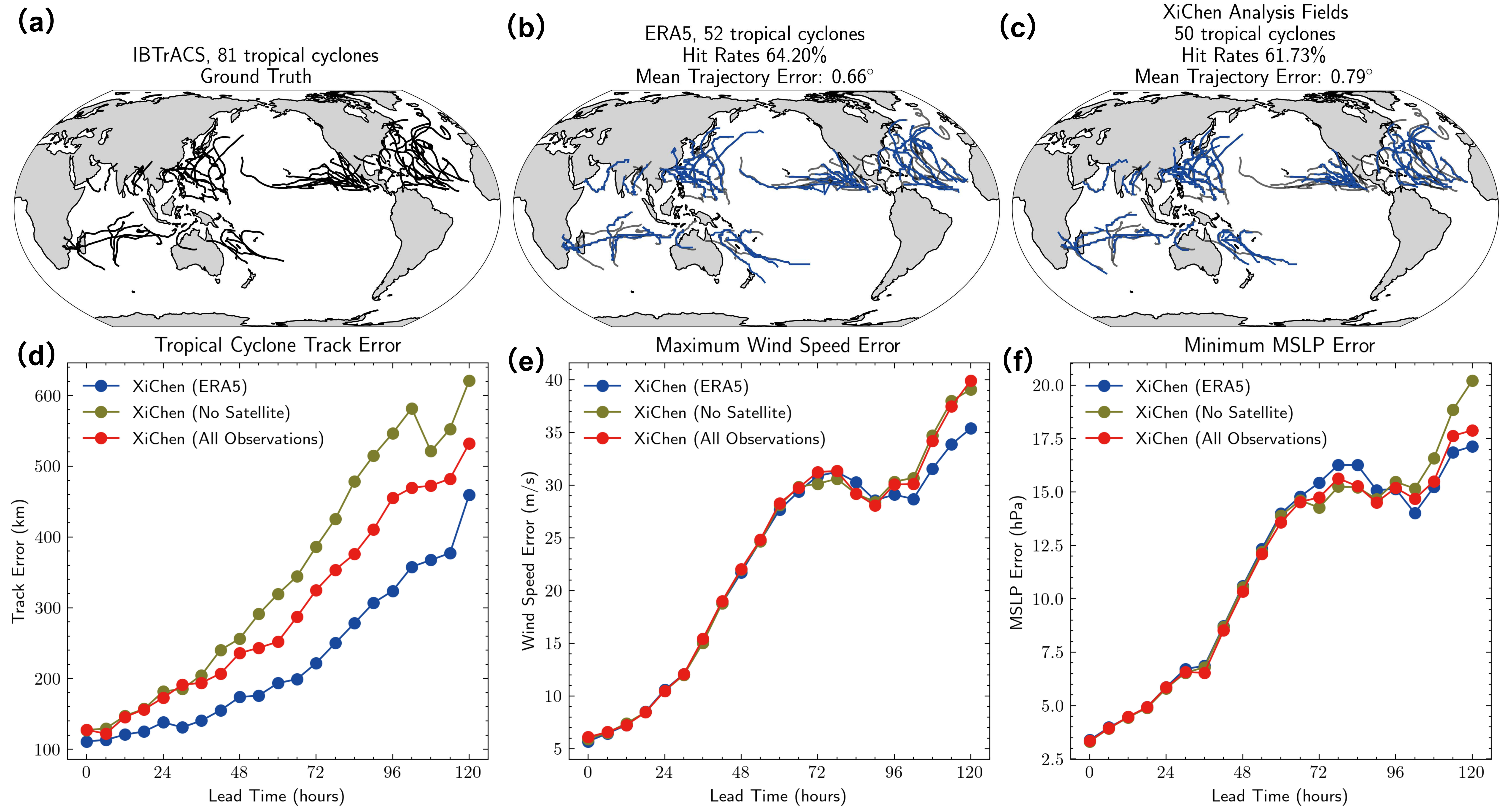}
    \caption{\textbf{Assimilation and Forecasting of TCs.} (a) TC trajectory tracks from IBTrACS ground truth data for the entire year of 2023. (b) TC trajectory tracks (blue lines) from the ERA5 reanalysis for 2023. (c) TC trajectory tracks (blue lines) from the XiChen analysis fields obtained by assimilating all observations during a one-year DA cycle. (d) Comparison of XiChen's performance in terms of mean direct position error over 5-day TC forecasts in 2023. (e) Comparison of XiChen's performance in terms of average maximum wind speed error over 5-day TC forecasts in 2023. (f) Comparison of XiChen's performance in terms of average minimum MSLP error over 5-day TC forecasts in 2023. In panels (d) through (f), the blue lines denote the outcomes initialized with ERA5 data, whereas the yellow and red lines correspond to results derived from analysis fields obtained by assimilating GDAS prepbufr exclusively and by further incorporating all satellite observations, respectively.}
\label{fig6}
\end{figure}
\FloatBarrier

Figures~\ref{fig6}(a)-(c) provide a comparative analysis of TC tracks from different datasets and XiChen's analysis fields. Figure~\ref{fig6}(a) depicts the TC trajectories from the IBTrACS ground truth dataset for 2023, including 81 TCs. The blue lines in Figure~\ref{fig6}(b) depict the TC trajectories identified in the ERA5 reanalysis dataset for 2023 that correspond with TC trajectories in IBTrACS, comprising a total of 52 tropical cyclones. The ERA5 dataset successfully identified 64.2\% of the TCs documented in IBTrACS, with an average trajectory deviation of 0.66 degrees. The blue lines in Figure~\ref{fig6}(c) represent TC trajectories generated by XiChen during the one-year DA cycle. The XiChen analysis fields successfully identified 61.73\% of the TCs documented in IBTrACS, with an average trajectory deviation of 0.79 degrees. These results demonstrate that XiChen effectively leverages multi-source observations to retain TC information within the analysis field. Notably, the TC trajectories generated by XiChen include some TCs not identified in ERA5, suggesting that XiChen not only reproduces ERA5 data but also exhibits a degree of generalization capability. As lead time increases, the number of TC tracks identified by XiChen decreases to a certain extent. This is primarily due to the smoothing effect inherent in ML-based weather forecasting (see Figure S11 for details).

To quantitatively evaluate the impact of satellite observations on track and intensity errors in XiChen's TC forecasts, we conducted 5-day predictions starting from the onset times of all typhoons in the IBTrACS dataset. Tracking and error calculations were performed at 6-hour intervals. The typhoon tracking methodology follows the scheme established by the Pangu-Weather system~\cite{bi2023accurate}. The sensitivity of tropical cyclone forecasts to different initial fields is examined using three configurations: ERA5 data, assimilation of GDAS prepbufr observations alone, and joint assimilation of GDAS prepbufr with all satellite observations (Figures~\ref{fig6}(d)–(f)). The inclusion of satellite observations leads to a noticeable reduction in TC trajectories forecast errors relative to using GDAS prepbufr alone (Figure~\ref{fig6}(d)), suggesting that satellite data provide additional dynamical constraints beneficial for TC trajectory prediction. Forecasts initialized from ERA5 nonetheless exhibit the highest overall skill, which is likely related to the fact that the weather forecasting model was trained solely on ERA5-based states. For intensity-related metrics, including maximum wind speed and minimum MSLP, differences among the three initialization strategies are relatively small (Figures~\ref{fig6}(e) and (f)). One possible explanation is that systematic biases in the TC intensity estimation of ERA5 data are inherited by XiChen during training, thereby limiting its sensitivity to alternative initial fields. In addition, many existing ML models tend to produce overly smoothed representations of atmospheric states, which can further hinder the accurate prediction of extreme intensity values~\cite{lam2023learning,fudanchen2023fuxi}. Overcoming these limitations remains an open problem for future work.

\section{Discussion}
Advances in weather forecasting rely on the integration of new observations, and the rapid development of ML with the increasing availability of atmospheric observations offers a new opportunity to revolutionize weather forecasting. In this study, we construct XiChen, a global weather data-to-forecast ML system via 4DVar gradient-guided observation-flexible assimilation. We demonstrate that the gradient of the 4DVar cost function provides a unified information interface for diverse heterogeneous observations, which enables physically consistent and observation-flexible DA for global weather forecasting. We verified that XiChen's assimilation incorporates physical equilibrium constraints and flow-dependent characteristics through single-point perturbation DA experiments. By assimilating GDAS prepbufr, MHS, ASCAT, AMSU-A, and SATWND observations, XiChen offers global weather observation-to-forecast capabilities and demonstrates superior performance compared to the GFS in both DA and forecasting tasks for several key variables. Notably, XiChen achieves a skillful lead time of over 8.75 days, which surpasses that of the GFS. Beyond weather forecasting performance, this work illustrates how physics-informed ML can serve as a bridge between heterogeneous Earth observations and data-driven models, offering a scalable pathway toward next-generation Earth system prediction.

Moreover, XiChen demonstrates exceptional flexibility by streamlining the assimilation of dynamic observation sources. When incorporating a new data source, the system only requires fine-tuning the foundation model to function as the respective observation operator, followed by fine-tuning the corresponding DA model within the proposed cascaded DA framework. Conversely, if an observation source becomes temporarily unavailable, the corresponding model can be excluded from the pipeline without necessitating a full-system retraining. These capabilities are critical for advancing ML-based DA and forecasting toward practical operationalization. This achievement highlights the growing role of AI in advancing operational weather forecasting and establishes XiChen as a foundation for leveraging data-driven approaches to better understand and predict Earth system dynamics with broad societal and environmental relevance. Rather than serving as independent alternatives, ML-based systems such as XiChen are best characterized as sophisticated extensions of existing NWP infrastructure. This complementary relationship stems from their fundamental reliance on NWP-generated reanalysis fields for training, positioning ML models as a vital augmentation of physics-based forecasting.

Although XiChen demonstrates significant potential for practical applications, it also exhibits certain limitations. First, the model experiences smoothing effects during both the DA and weather forecasting processes, a common challenge in ML-based weather forecasting models~\cite{bonavita2024some}. Potential solutions include designing more suitable loss functions~\cite{xu2024extremecast}, incorporating physical constraints~\cite{xu2024weathergft}, or employing post-processing techniques based on generative models~\cite{zhong2024fuxiextreme}. Second, the current implementation of XiChen integrates satellite observations from only four satellite instruments (MHS, ASCAT, AMSU-A, and SATWND), a limited subset compared to the extensive data streams utilized in operational NWP systems~\cite{zhong2024fuxiextreme}. In the future, incorporating additional indirect observations could significantly enhance the performance of the analysis field. Third, the model produces only deterministic weather forecasts. However, making critical decisions requires awareness of all possible weather scenarios and their associated probabilities. Therefore, it is necessary to extend the model to a probabilistic version. Addressing this limitation in the future may involve reconstructing the model to adopt a probabilistic framework~\cite{price2025probabilistic} or incorporating perturbations into the XiChen system~\cite{chen2024machine}. Finally, while XiChen demonstrates impressive performance in DA and medium-range global weather forecasting tasks, its deployment in operational settings would require addressing additional challenges related to real-time data ingestion, model updates, and integration with existing NWP infrastructure.

\section{Method}\label{sec-method}
\subsection{Notation and problem statement}
We denote the full state of the atmosphere at a particular time $t$ as a tensor $\mathbf{x}(t)$ with dimensions $V \times H \times W$, where $V$ represents the total number of variables, $H$ indicates the number of latitude coordinates, and $W$ signifies the number of longitude coordinates. The indexing scheme $\mathbf{x}_{v,i,j}(t)$ refers to the state of variable $v$ at time $t$, and the latitude-longitude coordinates $(i,j)$. Additionally, we represent GDAS prepbufr conventional atmospheric observations at time $t$ as a tensor $\mathbf{y}^c(t)$ of the same dimensions $V \times H \times W$. For unobserved variable $v$ at coordinates $(i,j)$, we set $\mathbf{y}^c_{v,i,j}(t)=\text{NAN}$, while for observed variable $v$, $\mathbf{y}^c_{v,i,j}(t)=\mathbf{x}_{v,i,j}(t)+\mathbf{\varepsilon}_{v,i,j}$, where $\mathbf{\varepsilon}_{v,i,j}$ represents the observation error. For satellite observations, we denote this as $\mathbf{y}^s(t)=\mathcal{H}^s\left(\mathbf{x}(t)\right)+\mathbf{\varepsilon}^s$, where $\mathbf{\varepsilon}^s$  is the satellite observation error and $\mathcal{H}^s$ is the observation operator that maps atmospheric variables to satellite observation variables. 

We aim to develop a global weather observation-to-forecast ML system that integrates both forecasting and DA models. For the forecasting model, given a system state $\mathbf{x}(t)$, our objective is to predict the state $\mathbf{x}(t + \Delta t)$ at a future time $t+\Delta t$. To accomplish this, the forecasting model $\mathcal{M}_{t \rightarrow t+\Delta t}: \mathbf{x}(t) \mapsto \hat{\mathbf{x}}(t+\Delta t)$ is trained to map the state of the atmosphere at present to a predicted state $\hat{\mathbf{x}}(t+\Delta t)$ for the future time $t + \Delta t$. Regarding the assimilation model, given a background field $\mathbf{x}^b(t_0)$ at time $t_0$, GDAS prepbufr observations $\{\mathbf{y}^c(t_0),\cdots,\mathbf{y}^c(t_K)\}$, and satellite observations $\{\mathbf{y}^s(t_0),\cdots,\mathbf{y}^s(t_K)\}$ during the DAW $[t_0,\cdots,t_K]$, our goal is to learn to estimate the optimal initial field $\mathbf{x}^a(t_0)$ by fusing these data. To assimilate the satellite observations, we need to learn an observation operator  $\hat{\mathcal{H}}^s: \left(\mathbf{x}(t),\theta^s(t)\right) \mapsto \mathbf{y}^s(t)$, where $\theta^s(t)$ denotes the auxiliary observational information at time $t$. Subsequently, we need to learn a DA model $\mathcal{D}: \left(\mathbf{x}^b(t_0),\{\mathbf{y}^c(t_0),\cdots,\mathbf{y}^c(t_K)\},\{\mathbf{y}^s(t_0),\cdots,\mathbf{y}^s(t_K)\}\right) \mapsto \mathbf{x}^a(t_0)$, where the reanalysis serves as the reference.

\subsection{Dataset}
In this study, we utilized the DABench~\cite{wang2024benchmark} dataset while incorporating MHS, ASCAT, AMSU-A, and SATWND satellite observations to train and evaluate the potential of XiChen for satellite assimilation tasks. DABench comprises 14 years of ERA5 reanalysis data, spanning from 2010 to 2023, as well as GDAS prepbufr observations, which were interpolated to a grid resolution of 1.40625$^\circ$ ($128 \times 256$ grids). The observations are temporally interpolated to achieve a 3-hour temporal resolution interval, whereby measurements are systematically acquired at 0, 3, 6, 9, 12, 15, 18, and 21 hours UTC daily. Observations obtained within a 1.5-hour temporal window preceding and succeeding these designated temporal reference points are considered temporally coincident. We examined 5 atmospheric variables (each with 9 pressure levels) and four surface variables, resulting in a total of 49 variables. The atmospheric variables include geopotential (Z), temperature (T), specific humidity (Q), the zonal component of wind (U), and the meridional component of wind (V). The 9 sub-variables at different vertical levels are denoted by abbreviating their short names along with their respective pressure levels (e.g., Z500 indicates the geopotential at 500 hPa). The 4 surface variables consist of 2-meter temperature (T2M), 10-meter zonal component of wind (U10), 10-meter meridional component of wind (V10), and mean sea level pressure (MSLP). For further details on the dataset, we refer readers to the article~\cite{wang2024benchmark}. 

For the AMSU-A and MHS observations, nearest neighbor interpolation was utilized to map both the raw satellite data and the auxiliary information onto a 1.40625$^\circ$ spatial resolution and 3-hour temporal resolution interval. In this study, we excluded observations from regions located above 60$^\circ$ north and south latitude to avoid complications due to sea ice ~\cite{noh2023assimilation}. Additionally, we implemented a straightforward quality control procedure that eliminated data with brightness temperature values exceeding 350 K or falling below 150 K. Since MHS observations are highly influenced by clouds, a basic cloud-detection method was applied to exclude cloud-contaminated portions of the MHS data. For details on the cloud-detection methodology, please refer to the reference paper~\cite{ma2013introduction}. For the 10-meter wind field data obtained from ASCAT and the upper-air wind field data obtained from SATWND observations, we adopt an identical processing methodology to that implemented for GDAS prepbufr observational datasets.

All models are trained on data collected until 2021, with the year 2022 designated as a validation set. Subsequently, data from 2023 is utilized for testing. For details of the reanalysis dataset as well as observations used in this study, see Supplementary Text and Tables S1 to S4.

\subsection{The ML-based global weather assimilation and medium-range forecasting system}~\label{sec-model}
The XiChen model is based on our proposed Conditional Hybrid Neural Operator (CHNO), which incorporates cross-attention to integrate conditional information into our Hybrid Neural Operator (HNO). This design enables a unified model architecture adaptable to multiple downstream tasks. Additionally, since the Adaptive Fourier Neural Operator (AFNO)~\cite{guibas2021adaptive} performs only global feature extraction through convolution operations in the Fourier space, we introduce a convolution-based branch to enhance local feature extraction capabilities, thereby developing the HNO. For further details on the CHNO and XiChen model architectures, refer to Figure S2. A comparison between HNO and AFNO in the context of medium-range global weather forecasting is provided in the Supplementary Text and Figure S5. 

Our system uses XiChen to cover core components of the NWP pipeline, from observation operators and DA models to the weather forecasting model. As shown in Figure~\ref{fig1}(b), XiChen integrates background field data, conventional observations, and satellite observations within the DAW to perform the DA task. The process begins by computing the 4DVar cost function, which measures the misfit between observations and model forecasts. The 4DVar cost function $\mathcal{J}^{\text{4DVar}}(\mathbf{x}(t_0))$ is defined as follows:
\begin{equation}
    \mathcal{J}^\text{4DVar}(\mathbf{x}^b(t_0)) = \frac{1}{2}\|\mathbf{x}^b(t_0)-\mathbf{x}(t_0)\|^2_{\mathbf{B}^{-1}}+\frac{1}{2}\sum^K_{k=0}\|\mathbf{y}(t_k)-\mathcal{H}\left(\mathcal{M}_{t_0 \rightarrow t_k}(\mathbf{x}(t_0))\right)\|^2_{\mathbf{R}^{-1}},
\end{equation}
where $\mathbf{x}(t_0)$ denotes the initial field to be optimized, $\mathbf{x}^b(t_0)$ represents the background field, and $\mathbf{y}(t_k)$ represents the observations at time $t_k$. The forecasting model, denoted as $\mathcal{M}_{t_0 \rightarrow t_k}$, maps the initial field to the field at time $t_k$. Additionally, $\mathbf{B}$ and $\mathbf{R}$ represent the background and observation error covariance matrices, respectively. The gradient of the 4DVar cost function with respect to the background field, denoted as $\nabla_{\mathbf{x}^b(t_0)}\mathcal{J}^{\text{4DVar}}(\mathbf{x}^b(t_0))$, is computed via automatic differentiation. Computing the gradient of the 4DVar cost function projects observational information onto the model state space, quantifying the necessary adjustments to the background field. Consequently, this process serves as a unified information interface for heterogeneous observations, obviating the need to design specialized encoders for distinct data types.

By combining the gradient $\nabla_{\mathbf{x}^b(t_0)}\mathcal{J}^{\text{4DVar}}(\mathbf{x}^b(t_0))$ with the background field $\mathbf{x}^b(t_0)$, XiChen generates the analysis field. Utilizing the analysis field along with the lead time as inputs, XiChen produces either a forecast field or an updated background field. Among them, our DA task involves sequentially employing DA models for different observational data. For example, in this study, XiChen first assimilates GDAS prepbufr data, then assimilates MHS, ASCAT, AMSU-A, and SATWND observations. This configuration enhances XiChen's flexibility. The assimilation of new observations in the future merely requires fine-tuning the respective observation operators and DA model components, eliminating the need to retrain the entire DA system. For details of the cascade DA framework, please refer to the Supplementary Text and Figure S1 (b).

\subsection{Pre-training for medium-range global weather forecasting}~\label{sec-pretrain}
We aim to design a foundation model that can be pre-trained on the reanalysis data and then be fine-tuned to solve various downstream weather tasks to build a global weather observation-to-forecast ML system. As illustrated in Figure~\ref{fig1}(c), during the pre-training process, the foundation model takes the reanalysis field $\mathbf{x}(t)$ as the input and the lead time $\Delta t$ as the condition and outputs the predicted state $\hat{\mathbf{x}}(t+\Delta t)$. In this study, we sample $\Delta t$ from the set $\{1, 3, 6, 12, 24\}$ hours. This process enables the foundation model to effectively capture the relationships among various weather variables while accounting for both the dynamical and thermodynamic characteristics of the atmosphere. To generate predictions for any time increments, the predictions from the learned forecasting model can be combined using the hierarchical temporal aggregation methodology proposed by Pangu-Weather~\cite{bi2023accurate}. Thus, the foundation model is pre-trained to minimize the following loss:
\begin{equation}
    \tiny{\mathcal{L}(\theta) = \mathbb{E}\left[\frac{1}{VHW}\sum^V_{v=1}\sum^H_{i=1}\sum^W_{j=1}\omega(v)L(i)\|\Delta \mathbf{x}_{v,i,j}(t+\Delta t)\|_1\right],}
\end{equation}
where $\omega(v)$ denotes the pressure-weighting coefficient of variable $v$~\cite{lam2023learning}, $\Delta \mathbf{x}(t+\Delta t) = \mathcal{M}_{t \rightarrow t+\Delta t}(\mathbf{x}(t))_{v,i,j}-\mathbf{x}_{v,i,j}(t+\Delta t)$ is the difference between the predicted state and the ground truth at time $t+\Delta t$, $\|\|_1$ represents the $L_1$-norm, and $L(i)$ is the latitude-weighting factor~\cite{rasp2020weatherbench}:
\begin{equation}
    L(i) = \frac{\cos\left(\text{lat}(i)\right)}{\frac{1}{H}\sum^H_{i'=1}\cos\left(\text{lat}(i')\right)},
\end{equation}
where $\text{lat}(i)$ is the latitude of the $i$th row of the grid. This term is commonly used in the training process of previous ML-based weather forecasting models~\cite{kurth2023fourcastnet,bi2023accurate,lam2023learning}. This pre-training process costs about 45 hours to run on 8 A100-40GB GPUs.

To mitigate accumulated errors in the forecasting process, we fine-tune the pre-trained forecasting model by employing a roll-out strategy for multi-step forecasting. Specifically, we execute the model $T$ times for each batch and calculate the average loss of these $T$-step forecasts to optimize the model parameters. The loss function for the multi-step forecast is as follows:
\begin{equation}
    \tiny{\mathcal{L}(\theta) = \mathbb{E}\left[\frac{1}{TVHW}\sum^T_{\tau=1}\sum^V_{v=1}\sum^H_{h=1}\sum^W_{w=1}\omega(v)L(i)\|\Delta \mathbf{x}_{v,i,j}(t+\tau\Delta t)\|_1\right],}
\end{equation}
where $\tau$ denotes the roll-out step and $\Delta \mathbf{x}_{v,i,j}(t+\tau\Delta t) = \mathcal{M}_{t \rightarrow t+\tau\Delta t}(\mathbf{x}(t))_{v,i,j}-\mathbf{x}_{v,i,j}(t+\tau\Delta t)$. 

In practice, we first fine-tune the pre-trained forecasting model with roll-out steps from $T=2$ to $T=4$ to develop the XiChen-Short model. The weights obtained from the XiChen-Short model are then used to initialize the XiChen-Medium model, which is subsequently fine-tuned to achieve optimal forecasting performance for periods of 5 to 10 roll-out steps. We use the same sampled value of $\Delta t$ for all $T$ steps. Our forecasting model is similar to FuXi~\cite{fudanchen2023fuxi} in its forecasting process. Since the analysis field errors provided by the XiChen DA model are comparable to those of XiChen-Short's 2-day forecasts initialized with ERA5, the 10-day medium-range global weather forecasts using initial fields estimated by the XiChen DA model are generated by XiChen-Short for the first 3 days and by XiChen-Medium for the remaining 7 days. The fine-tuning process of the XiChen forecasting model costs about 94 hours to run on 8 A100-40GB GPUs. For additional information on the model training configuration, please refer to the Supplementary Text. 

\subsection{Fine-tuning for the observation operator}\label{sec-obsop}
To compare the atmospheric state with satellite observations, the observation operator is essential~\cite{liang2023machine}. The simulated radiance is computed at each observation location using a profile of atmospheric variables. As illustrated in Figure~\ref{fig1}(d), we fine-tune the pre-trained model by taking the satellite's attitude parameters and scan position as conditions to construct an observation operator $\hat{\mathcal{H}}^s: \left(\mathbf{x}(t),\theta^s(t)\right) \mapsto \mathbf{y}^s(t)$, where $\theta^s(t)$ denotes the auxiliary observational information at time $t$. The loss function used to train $\hat{\mathcal{H}}^s$ is as follows:
\begin{equation}
    \mathcal{L} = \mathbb{E}\left[\frac{1}{C}\sum^C_{c=1}\|\hat{\mathcal{H}}^s\left(\mathbf{x}(t),\theta^s(t)\right)_c-\mathbf{y}^s_c(t)\|_1\right],
\end{equation}
where $C$ denotes the channels of the satellite observation, and the subscript $c$ represents the channel to be calculated. This process facilitates the development of scalable plug-ins to assimilate the satellite's raw observations. 

This study selected channels 5 to 10 of AMSU-A and 3 to 5 of MHS for training observation operators and conducting assimilation experiments. This selection was motivated by the fact that channels 5 to 10 of AMSU-A are sensitive to atmospheric temperature profiles ranging from 50 hPa to 1000 hPa, while channels 3 to 5 of MHS are responsive to the atmospheric water vapour in the troposphere~\cite{zou2017impacts}. The fine-tuning process of each observation operator costs about 13.3 hours to run on 8 A100-40GB GPUs. Please refer to the Supplementary Text for details regarding the configuration of the AMSU-A and MHS observation operators and their performance in Tables S7 and S8.

\subsection{Fine-tuning for data assimilation}\label{sec-ft-da}

As illustrated in Figure~\ref{fig1}(e), we fine-tune the pre-trained model to function as a DA model by incorporating the gradient of the 4DVar cost function with respect to the background field, denoted as $\nabla_{\mathbf{x}^b(t_0)}\mathcal{J}^{\text{4DVar}}(\mathbf{x}^b(t_0))$. This approach follows the derivation presented in 4DVarFormer~\cite{wang2024accurate}. However, unlike 4DVarFormer, XiChen outputs the representation of analysis increments in latent space rather than the analysis increments themselves. This representation is added to the embedding of the background field $\mathbf{x}^b(t_0)$ and subsequently decoded by the HNO blocks to generate the analysis field. 

In our 4DVar cost function, the observation covariance matrix of prepbufr observations is denoted as $\mathbf{R}^c = \varepsilon^c(\varepsilon^c)^T$. The observation operator error $\hat{\mathbf{\varepsilon}}^s$ is used to construct the error covariance matrix, $\mathbf{R}^s = \hat{\varepsilon}^s(\hat{\varepsilon}^s)^T$, of the satellite observations. Since we only need to compute the gradient of the initial value of the 4DVar cost function with respect to $\mathbf{x}^b$, there is no need to estimate $\mathbf{B}$. This simplifies the process and circumvents a challenging issue typically encountered in the traditional 4DVar method~\cite{huang2024adaptive}. In addition, unlike methods used in Aardvark~\cite{allen2025end}, GraphDOP~\cite{alexe2024graphdop}, FuXi-DA~\cite{xu2024fuxida}, and FengWu-Adas~\cite{chen2023fengwuadas}, we incorporate a quality control process into the computation of the 4DVar cost function by excluding observations whose distances from the background field exceed 5 times the standard deviation of the variable. This approach can prevent excessive adjustment of localized regions in the background field caused by large discrepancies between observations and the background. For a detailed depiction of the computational flow of the 4DVar cost function, please refer to Figure S3.

Therefore, we fine-tune the pre-trained model by taking $\nabla_{\mathbf{x}^b(t_0)}\mathcal{J}^{\text{4DVar}}(\mathbf{x}^b(t_0))$ as the input and the background field $\mathbf{x}^b$ as the condition to construct an DA model $\mathcal{D}$. The model is trained to minimize the following loss function:
\begin{equation}
    \mathcal{L}(\theta) = \mathbb{E}\left[\frac{1}{VHW}\sum^V_{v=1}\sum^H_{i=1}\sum^W_{j=1}\omega(v)L(i)\|\left(\mathbf{x}^a_{v,i,j}(t)-\mathbf{x}_{v,i,j}(t)\right)\|_1\right].
\end{equation}

We initially fine-tuned the pre-trained model using GDAS prepbufr observations to develop a prepbufr DA model, which costs about 45 hours to run on 8 A100-40GB GPUs. Subsequently, the prepbufr DA model was further fine-tuned into an MHS DA model, ASCAT DA model, AMSU-A DA model, and SATWND DA model. Detailed configurations for the assimilation model training are provided in the Supplementary Text.

To circumvent the temporal latency inherent in ML-based 4DVar methods~\cite{xiao2024fengwu4dvar,wang2024accurate} when deploying XiChen as a self-contained forecasting system for acquiring medium-range prognostic results, we have architected a dual assimilation framework (see the Supplementary Text and Figure S4 for details). This dual DA framework facilitates the real-time acquisition of medium-range global weather forecasting results without necessitating the temporal delay associated with the availability of all observations within the complete 12-hour DAW. Figures S9 and S10 illustrate the impact of varying assimilation windows on both DA cycle performance and 10-day medium-range forecast performance.

\subsection{Verification metrics}~\label{sec-metrics}
This study aims to assess our model's performance concerning the tasks of assimilation and forecasting. Consequently, this study comprehensively evaluates a one-year DA cycle and a 10-day medium-range forecast. The DA cycle is executed at 12-hour intervals for assimilations, with a DAW of 12 hours. Specifically, the DA cycle is run for the year 2023 at 00:00 UTC and 12:00 UTC each day, which corresponds to the initialization times for the 10-day forecast conducted by the IFS HRES, Pangu-Weather~\cite{bi2023accurate}, GraphCast~\cite{lam2023learning}, and FengWu~\cite{ailabchen2025fengwu}. To evaluate medium-range forecasts, we follow the WeatherBench~\cite{rasp2020weatherbench} and DABench~\cite{wang2024benchmark}, selecting 50 initial fields at 336-hour intervals for the medium-range forecast experiments. The first initial field at 00:00 UTC is set for January 1, 2023, while the first initial field at 12:00 UTC is on January 8, 2023.

All metrics were computed using float32 precision and reported using the native scale of the variables without normalization. Notably, all metrics are computed using a latitude-weighting factor over grid points due to the non-equal area distribution from the equator towards the north and south poles. 

\paragraph{Root Mean Square Error (RMSE)} We evaluate assimilation and forecast skills for a given variable, $\mathbf{x}_v$, using a latitude-weighted Root Mean Square Error (RMSE)~\cite{rasp2020weatherbench} given by
\begin{equation}
    \text{RMSE}_v = \frac{1}{|D_{eval}|}\sum_{d \in D_{eval}}\sqrt{\frac{1}{HW}\sum^{H}_{i=1}\sum^{W}_{j=1} L(i)(\hat{\mathbf{x}}_{v,i,j}-\mathbf{x}^t_{v,i,j})^2},
\end{equation}
where 
\begin{itemize}
    \item $\hat{\mathbf{x}}$ is the field to be evaluated,
    \item $\mathbf{x}^t$ is the ERA5 ground truth,
    \item $d \in D_{eval}$ represent the sample index in the evaluation dataset,
    \item $i$ represents the latitude coordinate in the grid,
    \item $j$ represents the longitude coordinate in the grid.
\end{itemize}
The lower the RMSE, the better the results.

\paragraph{Anomaly Correlation Coefficient (ACC)} To study skillful forecast lead times, we also calculated the latitude-weighted Anomaly Correlation Coefficient (ACC)~\cite{rasp2020weatherbench} according to
\begin{equation}
    \text{ACC}_v = \frac{\sum_{d \in D_{eval}}\frac{\sum^H_i\sum^W_jL(i)\left(\hat{\mathbf{x}}_{v,i,j}-\mathbf{C}_{v,i,j}\right)\left(\mathbf{x}^t_{v,i,j}-\mathbf{C}_{v,i,j}\right)}{\sqrt{\left[\sum^H_i\sum^W_jL(i)\left(\hat{\mathbf{x}}_{i,j}-\mathbf{C}_{v,i,j}\right)^2\right]\left[\sum^H_i\sum^W_jL(i)\left(\mathbf{x}^t_{v,i,j}-\mathbf{C}_{v,i,j}\right)^2\right]}}}{|D_{eval}|},
\end{equation}
where $\mathbf{C}_{v,i,j}$ denotes the climatological mean for a given variable $v$ and the day-of-year containing the validity time at the ${i,j}$ grid point. It is calculated referring to GraphCast~\cite{lam2023learning}, FengWu~\cite{ailabchen2025fengwu}, and DABench~\cite{wang2024benchmark}. The hourly climatological fields are computed using ERA5 data between 2010 and 2021. The higher the ACC, the better the results.

\backmatter

\bmhead{Supplementary information}
The supplementary material is available at the ``Supplementary\_Information.pdf".

\bmhead{Acknowledgements}
The authors extend their gratitude to the ECMWF and NCEP for their significant efforts to store and provide invaluable data, which are crucial for this work and the research community. Additionally, this work was carried out at the National Supercomputer Center in Tianjin, and the calculations were performed on Tianhe new generation supercomputer. We would also like to express our appreciation to the research team and service team in the Shanghai Artificial Intelligence Laboratory for the provision of computational resources and infrastructure. % This work was supported by the Science and Technology Innovation Program of Hunan Province (2022RC3070) and the National Natural Science Foundation of China~(Grant Nos.\, 42205161, 42405146, and 42275170). 

\bmhead{Funding}
Not applicable. 

\bmhead{Conflict of interest}
All authors declare no financial or non-financial competing interests. 

\bmhead{Ethics approval and consent to participate}
Not applicable. 

\bmhead{Consent for publication}
Not applicable. 

\bmhead{Availability of data and materials}
All data needed to evaluate the conclusions in the paper are present in the paper and/or the Supplementary Information. The DABench dataset is available at \href{https://pan.baidu.com/s/1H3_H2Xvy8OAZXIL-W7woYw?pwd=pdvn}{the Baidu Drive}. The satellite observation MHS, ASCAT, AMSU-A, and SATWND observations can be accessed \href{https://rda.ucar.edu/datasets/d735000/}{here}. The TC ground truth used in this study is obtained from the IBTrACS dataset, accessible at \href{https://www.ncei.noaa.gov/products/international-best-track-archive}{here}, which compiles recent and historical TC records from multiple agencies and provides the most comprehensive global archive of TC information.

\bmhead{Code availability}
The source code used for this work is available in a GitHub repository \href{https://github.com/wuxinwang1997/XiChen_1.40625deg}{https://github.com/wuxinwang1997/XiChen\_1.40625deg}.

\bmhead{Author contribution}
K.J.R., L.B., B.H.D., H.Z.L., W.M.Z., J.Q.S., and W.X.W. designed the project. K.J.R., L.B., W.C.N., and B.H.D. managed and oversaw the project. W.X.W., S.M., and Y.L.Z. selected and preprocessed the observations. W.X.W., T.K.Y., and T.H. performed the model training and evaluation. W.X.W., T.H., and L.L.H. improved the model design. W.X.W., T.K.Y., X.Y.L., K.F.D., and T.H. analyzed the experimental results. W.X.W., B.F., and W.C.N. wrote and revised the manuscript.

\bibliography{sn-bibliography}% common bib file

\end{document}